\documentclass[letterpaper, 10 pt, conference]{ieeeconf}  % Comment this line out if you need 

\IEEEoverridecommandlockouts                              % This command is only needed if 
\usepackage{adjustbox}
\usepackage[ruled, vlined]{algorithm2e}
\usepackage{amsmath}
\usepackage{amssymb}
\usepackage{array}  %
\usepackage[accsupp]{axessibility}  %
\usepackage{booktabs}
\usepackage{capt-of}
\makeatletter
\let\NAT@parse\undefined
\makeatother
\usepackage{cite}
\usepackage[font={small}]{caption}
\usepackage{colortbl}
\usepackage{epsfig}
\usepackage{inconsolata}  %
\usepackage{graphicx}
\usepackage{listings}
\usepackage{makecell}
\usepackage{multicol}
\usepackage{multirow}
\usepackage{paralist}
\usepackage{pifont}
\usepackage{tabularray}
\usepackage{times}
\usepackage{tabularx}
\usepackage{svg}

\definecolor{flodarkpurple}{rgb}{0.288,0.1196,0.7}

\definecolor{amber}{rgb}{1.0, 0.75, 0.0}

\usepackage[backref=page,breaklinks,colorlinks,bookmarks,citecolor=flodarkpurple]{hyperref}
\usepackage{listings}
\usepackage{comment}
\usepackage{arydshln}
\usepackage{tabularx}
\usepackage{xcolor}
\usepackage{colortbl}
\usepackage{xcolor}
\definecolor{codegreen}{rgb}{0,0.6,0}
\definecolor{codegray}{rgb}{0.5,0.5,0.5}
\definecolor{codepurple}{rgb}{0.18,0,0.82}
\definecolor{backcolour}{rgb}{0.98,0.98,0.98}
\colorlet{numb}{magenta!60!black}
\definecolor{Gray}{gray}{0.85}
\newcolumntype{a}{>{\columncolor{Gray}}c}

\colorlet{punct}{red!60!black}
\definecolor{background}{HTML}{EEEEEE}
\definecolor{delim}{RGB}{20,105,176}
\colorlet{numb}{magenta!60!black}

\lstdefinelanguage{json}{
    basicstyle=\footnotesize\ttfamily,
    numbers=left,
    numberstyle=\footnotesize,
    stepnumber=1,
    identifierstyle=\color{magenta},
    numbersep=8pt,
    showstringspaces=false,
    breaklines=true,
    frame=lines,
    backgroundcolor=\color{background},
    literate=
     *{0}{{{\color{numb}0}}}{1}
      {1}{{{\color{numb}1}}}{1}
      {2}{{{\color{numb}2}}}{1}
      {3}{{{\color{numb}3}}}{1}
      {4}{{{\color{numb}4}}}{1}
      {5}{{{\color{numb}5}}}{1}
      {6}{{{\color{numb}6}}}{1}
      {7}{{{\color{numb}7}}}{1}
      {8}{{{\color{numb}8}}}{1}
      {9}{{{\color{numb}9}}}{1}
      {:}{{{\color{punct}{:}}}}{1}
      {,}{{{\color{punct}{,}}}}{1}
      {\{}{{{\color{delim}{\{}}}}{1}
      {\}}{{{\color{delim}{\}}}}}{1}
      {[}{{{\color{delim}{[}}}}{1}
      {]}{{{\color{delim}{]}}}}{1},
}
\lstdefinestyle{mystyle}{
    backgroundcolor=\color{backcolour},   
    commentstyle=\color{codegreen},
    keywordstyle=\color{magenta},
    identifierstyle=\color{red},    
    numberstyle=\tiny\color{codegray},
    escapeinside={\%*}{*)},
    basicstyle=\ttfamily\footnotesize,
    breakatwhitespace=false,         
    breaklines=true,                 
    captionpos=b,                    
    keepspaces=true,                 
    numbersep=5pt,                  
    showspaces=false,                
    showstringspaces=false,
    showtabs=false,                  
    tabsize=2,
    literate=
     *{0}{{{\color{numb}0}}}{1}
      {1}{{{\color{numb}1}}}{1}
      {2}{{{\color{numb}2}}}{1}
      {3}{{{\color{numb}3}}}{1}
      {4}{{{\color{numb}4}}}{1}
      {5}{{{\color{numb}5}}}{1}
      {6}{{{\color{numb}6}}}{1}
      {7}{{{\color{numb}7}}}{1}
      {8}{{{\color{numb}8}}}{1}
      {9}{{{\color{numb}9}}}{1}    
}

\newcommand{\coolname}{\textit{Talk2BEV}}
\newcommand{\webpage}{https://llmbev.github.io/talk2bev/}

\newcommand{\authorhref}[3][flodarkpurple]{\href{#2}{\color{#1}{#3}}}

\title{\Large \bf \coolname: Language-enhanced Bird's-eye View Maps for Autonomous Driving\\
\vspace{0.30em}
\large{\href{\webpage}{\color{violet}{\texttt{\webpage}}}}
}

\author{
\authorhref{https://github.com/tusharc31/}{Tushar Choudhary}$^{1*}$,
\authorhref{https://vikr-182.github.io/}{Vikrant Dewangan}$^{1*}$, 
\authorhref{https://scholar.google.com/citations?user=ZER2BeIAAAAJ\&hl=en}{Shivam Chandhok}$^{2*}$, 
\authorhref{https://github.com/RudeNinja}{Shubham Priyadarshan}$^{1}$, 
\authorhref{}{Anushka Jain}$^{1}$, 
\\
\authorhref{https://scholar.google.co.in/citations?user=0zgDoIEAAAAJ\&hl=en}{Arun K. Singh}$^{3}$,
\authorhref{https://siddharthsrivastava.github.io/}{Siddharth Srivastava}$^{4}$,
\authorhref{https://krrish94.github.io/}{Krishna Murthy Jatavallabhula}$^{5\dagger}$, and 
\authorhref{https://scholar.google.co.in/citations?user=QDuPGHwAAAAJ\&hl=en}{K. Madhava Krishna}$^{1\dagger}$
\\
$^{1}$\href{https://robotics.iiit.ac.in/}{IIIT Hyderabad}, 
$^{2}$\href{https://www.ubc.ca/}{University of British Columbia}, 
$^{3}$\href{https://ut.ee/en}{University of Tartu},
$^{4}$\href{https://tensortour.com/}{TensorTour Inc.},
$^{5}$\href{https://www.csail.mit.edu/}{MIT}
\thanks{$^*$Equal contribution. \hspace{0.3cm} $^\dagger$ Equal advising. }% <-this % stops a space
}

\begin{document}

% Update Vspacing between title & author list
% Original (Line 3633 ieeeconf.cls): 
% \vskip0.25in{\LARGE\@title\par}\vskip1.0em\par
% Modified:
% \vskip0.25in{\LARGE\@title\par}\vskip0.2em\par

\makeatletter
\let\@oldmaketitle\@maketitle
\renewcommand{\@maketitle}{\@oldmaketitle
\centering
\includegraphics[width=\linewidth]{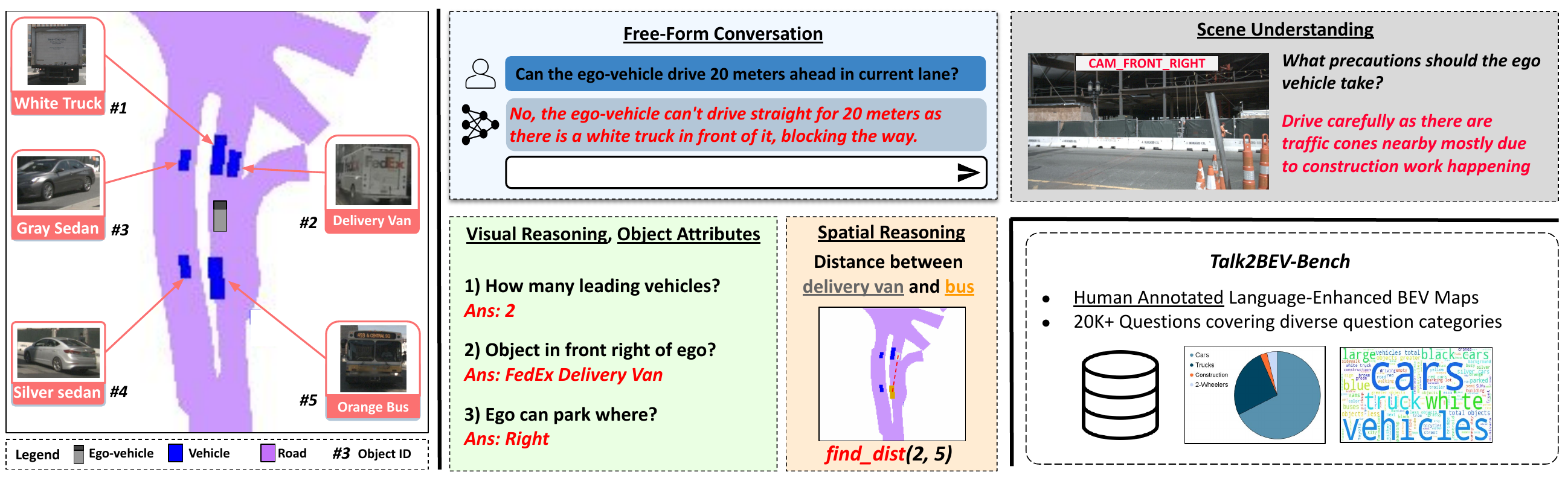}
\captionof{figure}{
\textbf{\coolname{}} builds \emph{language-enhanced bird's-eye view (BEV) maps} using (a) BEV representations constructed from vehicle sensors (Multi-View Images, LiDAR), and (b) Aligned vision-language features for each object which can be directly used as context within large vision-language models (LVLMs) to query and \emph{talk} to the objects in the scene. These maps embed knowledge about object semantics, material properties, affordances, and spatial concepts and can be queried for visual reasoning, spatial understanding, and making decisions about potential future scenarios, critical for autonomous driving application. Further, we develop the first benchmark \emph{Talk2BEV-Bench} towards evaluating LVLMs for AD applications spanning a diverse set of question categories with human-annotated ground-truth.
% which assesses aspects including,
% but not limited to object attributes, semantics, visual
% reasoning, spatial understanding, and decision-making.
}
\label{fig:splash}
}
\makeatother
%https://docs.google.com/drawings/d/15h7LI8KhEV6l9u_sVL2UQebIKBfWl4g92qv0MsCuh9E/edit?usp=sharing

\maketitle
\thispagestyle{empty}
\pagestyle{empty}

%%%%%%%%%%%%%%%%%%%%%%%%%%%%%%%%%%%%%%%%%%%%%%%%%%%%%%%%%%%%%%%%%%%%%%%%%%%%%%%%
\begin{abstract}
This work introduces \coolname{}, a large vision-language model (LVLM)\footnote{In this work, we use this term to refer to instruction-finetuned vision-language models; i.e., models that can consume text and image as input, and outupt text conditioned on the visual context~\cite{li2023blip2,minigpt4,llava}.} interface for bird's-eye view (BEV) maps in autonomous driving contexts.
While existing perception systems for autonomous driving scenarios have largely focused on a pre-defined (closed) set of object categories and driving scenarios, \coolname{} blends recent advances in general-purpose language and vision models with BEV-structured map representations, eliminating the need for task-specific models.
This enables a single system to cater to a variety of autonomous driving tasks encompassing visual and spatial reasoning, predicting the intents of traffic actors, and decision-making based on visual cues.
We extensively evaluate \coolname{} on a large number of scene understanding tasks that rely on \emph{both} the ability to interpret free-form natural language queries, and in grounding these queries to the visual context embedded into the language-enhanced BEV map.
To enable further research in LVLMs for autonomous driving scenarios,
we develop and release \textit{Talk2BEV-Bench}, a benchmark encompassing 1000 human-annotated BEV scenarios, with more than 20,000 questions and ground-truth responses from the NuScenes dataset.
We encourage the reader to view the demos on our project page: \url{https://llmbev.github.io/talk2bev/}
\end{abstract}

\setcounter{figure}{1} % Hotfix, for consistent figure numbers (else, after the teaser fig, latex skips to fig 3)

%%%%%%%%%%%%%%%%%%%%%%%%%%%%%%%%%%%%%%%%%%%%%%%%%%%%%%%%%%%%%%%%%%%%%%%%%%%%%%%%
\section{Introduction}
\label{sec:intro}

For safe navigation without human intervention, autonomous driving (AD) systems need to understand the visual world around them to make informed decisions.
This entails not just recognizing specific object categories, but also contextualizing their current and potential future interactions with the environment.
Existing AD systems rely on domain-specific models for each scene understanding task, such as detecting traffic actors and signage or forecasting plausible future events.
On the other hand, recent advances in large language models (LLMs) \cite{chung2022scaling, zheng2023judging, chatgpt2021, touvron2023llama, touvron2023llama2} and large vision-language models (LVLMs)\cite{openai2023gpt4, llava, minigpt4, dai2023instructblip} have demonstrated a promising alternative to thinking about perception for AD; that of a single model pretrained on web-scale data, capable of performing all the aforementioned tasks and more (particularly, the ability to deal with unforeseen scenarios).
In this work we ask, \emph{how do we most efficiently integrate such capabilities of LLMs with scene represenations traditionally used in autonomous driving}?

\begin{figure*}[!ht]
    \centering
    \includegraphics[width=\linewidth]{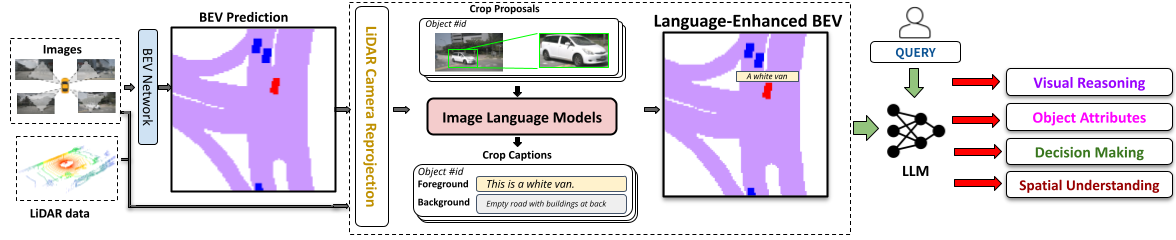}
    \caption{\textbf{Overall \coolname{}  Pipeline}: We first generate the bird's-eye view (BEV) map from image and LiDAR data. We then construct the language-enhanced map by augmenting the generated BEV with aligned image-language features for each object from large vision-language models (LVLMs). These features can directly be used as context to LVLMs for answering object-level and scene-level queries.
    For each object in the BEV, we project it to the image (using LiDAR-camera extrinsics), extract a bounding box, and caption the cropped bounding box using an off-the-shelf LVLM. Each object in the language-enhanced map now encodes geometric cues (position, area, centroid), and semantic cues (object and image descriptions).
    \vspace{-15pt}
    }
    \label{fig:open-set}
\end{figure*}

To this end, we introduce \coolname{}, language-enhanced maps for AD that enable holistic scene understanding and reasoning across a broad class of road scenarios.
Our framework interfaces LVLMs with bird's-eye view (BEV) maps---top-down semantic maps of the road plane and traffic actors that are widely used in AD systems \cite{philion2020lift, li2022bevformer, hu2021fiery, hu2022stp3} ---to enable visual reasoning, spatial understanding, and decision-making.
We augment a BEV map with aligned image-language features for each object in the scene.
These features can then directly be passed as (visual) context to an LVLM, enabling the model to answer a wide range of questions about the scene and make decisions about potential future scenarios using the vast knowledge base acquired by the LVLM during training.
We find that these LVLMs can interpret object semantics, material properties, affordances, and spatial concepts; and are an ideal alternative to domain-specific models.

Notably, our approach does not require any BEV-specific or vision-language training/finetuning; and uses existing pretrained LLMs and LVLMs.
This allows our approach to be flexibly and raplidly deployed on a wide class of domains and tasks, and to easily adapt to newer LLMs and LVLMs as newer and better models become available.

To objectively evaluate LVLMs for perception in the AD context and to expedite further reserch,  we also develop Talk2BEV-Bench:
a benchmark for the evaluation of large vision and language models for autonomous driving systems on a range of tasks, encompassing object-level and scene-level visual understanding.

In summary, our contributions are as follows
\begin{itemize}
    \item We develop \coolname{}, the first system to augment BEV maps with language to enable general-purpose visuolinguistic reasoning for AD scenarios.
    \item Our framework does not require any training or fine-tuning, relying instead on pre-trained image-language models. This allows generalization to a diverse collection of models, scenarios, and tasks.
    \item We develop Talk2BEV-Bench, a benchmark for evaluating LVLMs for AD applications with human-annotated ground-truth for object attributes, semantics, visual reasoning, spatial understanding, and decision-making. 
\end{itemize}

\section{Related work}
\label{sec:related_work}
\noindent\textbf{Large Vision Language Models.}
Recent advancements in large language models (LLMs)\cite{chung2022scaling, zheng2023judging, chatgpt2021, touvron2023llama, touvron2023llama2} and large vision-language models (LVLMs)\cite{openai2023gpt4, llava, minigpt4, dai2023instructblip} have emerged over the last few months. Evaluating and benchmarking these models remains challenging, with several proposals exploring LVLM benchmarking~\cite{fu2023mme,liu2023mmbench,xu2023lvlmehub} through the curation of question-response pairs using off-the-shelf LLMs, which compromises objectivity. Addressing this, SEEDBench~\cite{li2023seedbench} introduces a criterion where each question has four potential responses, and the LVLM ranks the best response. We adopt this evaluation methodology for its objectivity.

\noindent\textbf{3D Vision-Language Models.} 
LVLMs have also begun to be applied in scene understanding tasks such as object localization~\cite{Achlioptas2020ReferIt3DNL,Huang2021TextGuidedGN, feng2021freeform}, scene captioning \cite{chen2020scanrefer, chen2020scan2cap}, 3D Visual Question Answering utilizing multi-view images~\cite{azuma2022scanqa, chou2020visualqa} or point clouds~\cite{wijmans2019embodiedvqa, yan2021clevr3d}.
3D-LLM \cite{hong20233dllm} integrates LLMs into point clouds from multi-view images, bridging 2D models to 3D. In contrast, Point-LLM \cite{xu2023pointllm} trains solely on point clouds, bypassing the need for images.

\noindent\textbf{Vision-Language Models for Autonomous Driving.} 
Another recent trend, relevant to this work, is the adoption of LVLMs for autonomous driving~\cite{talk2thecariiit,talk2car,wu2023language}.
CityScapes-Ref~\cite{cityscapesref}, Talk2Car~\cite{talk2car} perform language-grounding tasks using the CityScapes~\cite{cordts2016cityscapes} and NuScenes~\cite{caesar2020nuscenes} datasets respectively.
ReferKITTI~\cite{wu2023referring} leverages temporal data for referring object detection and tracking on the KITTI dataset.
NuPrompt~\cite{wu2023language} leverages 3D pointcloud information using RoBERTa~\cite{liu2019roberta} as their language encoder.
Our work offers substantial improvements over this by blending state-of-the-art LLMs and LVLMs with BEV maps, while requiring no training or finetuning.

\noindent \textbf{Concurrent Work}: We briefly review recent and unpublished pre-prints that have surfaced after this work had been finalized.
NuScenes-QA~\cite{qian2023nuscenesqa} addresses Visual Question Answering (VQA) in autonomous driving by crafting scene graphs and question templates. Their evaluation demands end-to-end training and exact answer matching.
Other efforts have focused on training end-to-end vision-language-action models~\cite{lingo1} on large amoungs of aligned multimodal data.
In contrast to earlier methods, we offer zero-shot scene comprehension using LVLM's generalization and introde a broader benchmark, \textit{Talk2BEV-Bench}, to assess LVLMs for scene understanding via BEVs in autonomous driving.

\section{Talk2BEV}
\label{sec:approach}

\begin{figure*}
    \centering
    \includegraphics[width=\linewidth]{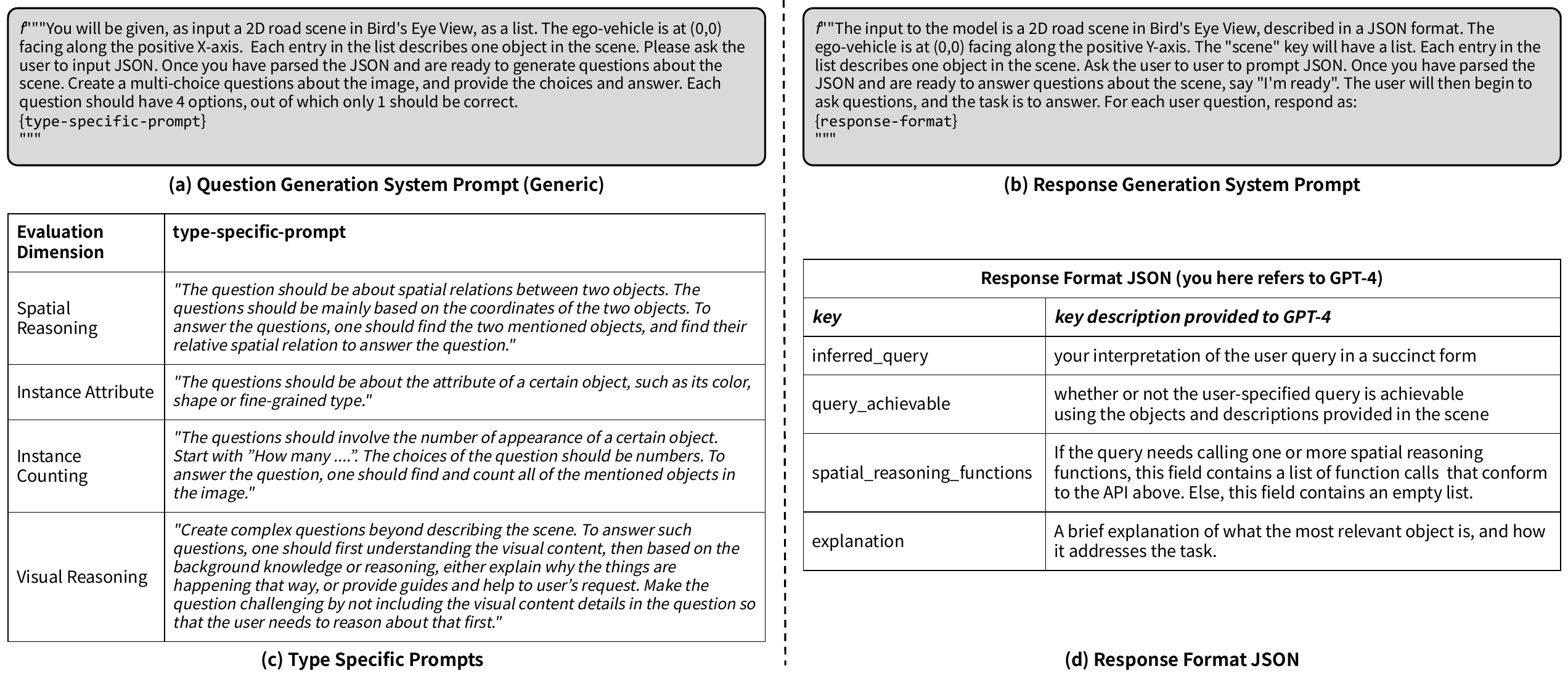}
    \caption{\textbf{LLM System Prompts:} \textbf{(a)} Generic question generation prompt for the LLM~\cite{openai2023gpt4}. \textbf{(b)} System prompt for response generation. \textbf{(c)} Details the type-specific commands added to generate questions along each evaluation dimension. \textbf{(d)} Displays the response format JSON with a brief explanation provided to LLM as to how it should fill each key of the JSON. }
    \label{fig:gpt4-prompts}
\end{figure*}

The key idea of \coolname{} is to enhance a birds-eye view (BEV) map with general-purpose vision-language features derived from pretrained LVLMs. A BEV map, denoted $\mathcal{O}$, is a top-view multi-channel grid encoding semantic information (in this work, only \textit{vehicle} and \textit{road})\footnote{We use the \textit{vehicle} class to extract LVLM features, and the \textit{road} class only for visualization purposes.}. The ego-vehicle is at the origin, assumed to be the center of the BEV. Given multi-view RGB images $\mathcal{I}$ a LiDAR pointcloud $\mathcal{X}$, a BEV can be obtained using a number of off-the-shelf approaches~\cite{philion2020lift,hu2022stp3,li2022bevformer,monolayout,autolay}.

Our three-phase pipeline (see Fig.~\ref{fig:open-set}) proceeds as follows:
\begin{enumerate}
    \item We first estimate a BEV map using onboard vehicle sensors (multi-view images) using an off-the-shelf BEV prediction model~\cite{philion2020lift}.
    \item For each object in this BEV map, we generate aligned image-language features using an LVLM~\cite{li2023blip2, dai2023instructblip, minigpt4}.
    These features are then passed into the language model of an LVLM to extract object metadata. 
    The object data, in conjunction with geometric information encapsulated in the BEV, forms the language-enhanced map, $\textbf{L}(\mathcal{O})$.
    \item Finally, given a user query, we prompt an LLM (eg. GPT-4~\cite{openai2023gpt4}) which interprets this query, parses the language-enhanced BEV as needed, and produces a response to this query.
\end{enumerate}

\subsection{Language Enhanced Maps}

\noindent\textbf{BEV-Image Correspondence.} First, we localize each object in the estimated BEV across the multi-view images used to produce the BEV map.
For each object in the BEV map, we compute a set of $k$ closest points in the LiDAR scan (a pointcloud); and project them into the camera frame using an inverse homography.

\noindent\textbf{Map Representation.} Our language-enhanced map augments the set of objects in a BEV by computing the image region corresponding to the object and deriving spatial and textual descriptions.
For each object $i$, we compute
(a) displacement along the BEV X and Y axes (in $m$) from the ego-vehicle, (b) object area (in $m^2$), (c) a text description of the object, and (d) a text description of the background.
LVLMs are specifically prompted to generated detailed descriptions of objects, and their outputs typically encode the type, color, and utility of the vehicle, status of the vehicle indicators, any text displayed on the vehicle, and more\footnote{All prompts we used are made available on our webpage.}.

\begin{figure}[!htbp]
    \centering
    \includegraphics[width=\linewidth]{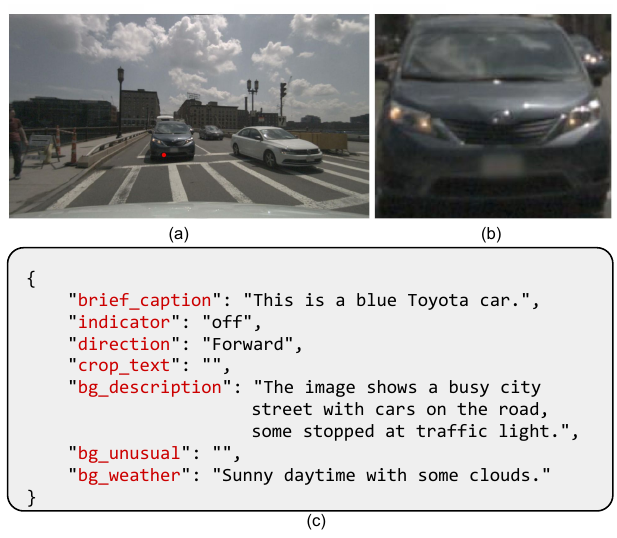}
    \caption{\textbf{Crop Description}: (a) A sample image $\mathbf{I}_n \in \mathcal{I}$ along with (b) the object crop $r_i$ and (c) its description $c_i$. }
    \label{fig:sample-crop-description}
\end{figure}

\noindent\textbf{Language-enhancement.} We then use a point-queryable segmentation model, such as FastSAM~\cite{fastsam} with a point-prompt (the center of the image crop) to generate instance segmentation masks.
The $k$ back-projected points serve as positive labels to the point-prompt.
For each segmentation mask, we crop a tight-fit bounding box and pass it to an LVLM to generate descriptions for the crop.
At this stage, we only pass the cropped bounding boxes through the visual encoders, to obtain image-language features that may later be passed as context tokens into language decoders.
The descriptions for each object encompass both object-level and scene-level details.
These generated metadata are then added to the BEV map in the form of a text entry (see sample JSON-structured entry below, and in Fig.~\ref{fig:sample-crop-description}).

\begin{lstlisting}[language=json, frame=single]
[...,
{   "object_id": 3,
    "position": [2.5, 1.5],
    "area": 4,
    "crop_descriptions": {...}
},...]
\end{lstlisting}     

\subsection{Response Generation}
\begin{figure*}[!ht]
    \centering
    \includegraphics[width=\linewidth]{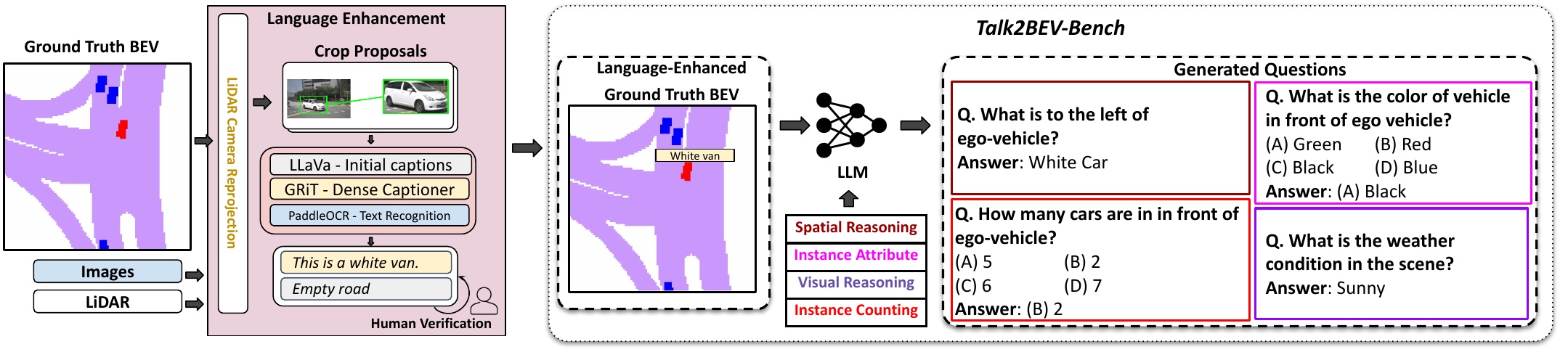}
    \caption{\textit{Talk2BEV-Bench} Creation: To
develop this benchmark, we use the NuScenes Ground Truth
BEV annotations and generate object and scene-level descriptions using dense Captioners (GRiT \cite{grit}), and Text-Recognition (PaddleOCR \cite{du2020ppocr}) models. The Ground Truth BEV is then passed to an LLM like GPT4 to generate diverse questions including, but not limited to- Spatial Reasoning, Instance Attribute, Visual Reasoning and Instance Counting.\vspace{-15pt}}
    \label{fig:json_generation}
\end{figure*}

\noindent\textbf{Type of queries.} 
The \textit{Talk2BEV} system can handle multiple kinds of user queries. In this work, we categorize them into free-form text queries, multiple choice questions (MCQ) with one correct answer, and spatial reasoning queries (specified via text).
Free-form and spatial reasoning queries emulate the natural end-user interface for \coolname{}, whereas MCQs allow us to perform objective evaluation, following the protocol outlined in SEEDBench~\cite{li2023seedbench}.

\noindent \textbf{Response format}: Opposed to directly producing free-form text outputs, we instruct the LLM used in \coolname{} to produce a JSON-formatted output with four fields (i) \texttt{inferred\_query}, which rephrases the user query first, thereby providing its internal interpretation of that query; (ii) \texttt{query\_achievable}, indicating whether or not the query is achievable. (iii) \texttt{spatial\_reasoning\_functions}, denoting whether spatial reasoning functions are needed, and (iv) \texttt{explanation}, containing a brief explanation of how the LLM addressed the provided task. Fig \ref{fig:gpt4-prompts} specifies the system prompts provided to LLM (GPT-4 in this case).
This format offers dual advantages: first, it ensures the LLM delivers information organized into key-value pairs. Second, it enables chain-of-thought reasoning~\cite{wei2023chainofthought} by outlining the intermediate steps that lead to the final response.

\begin{figure}[!htbp]
    \centering
    \includegraphics[width=\linewidth]{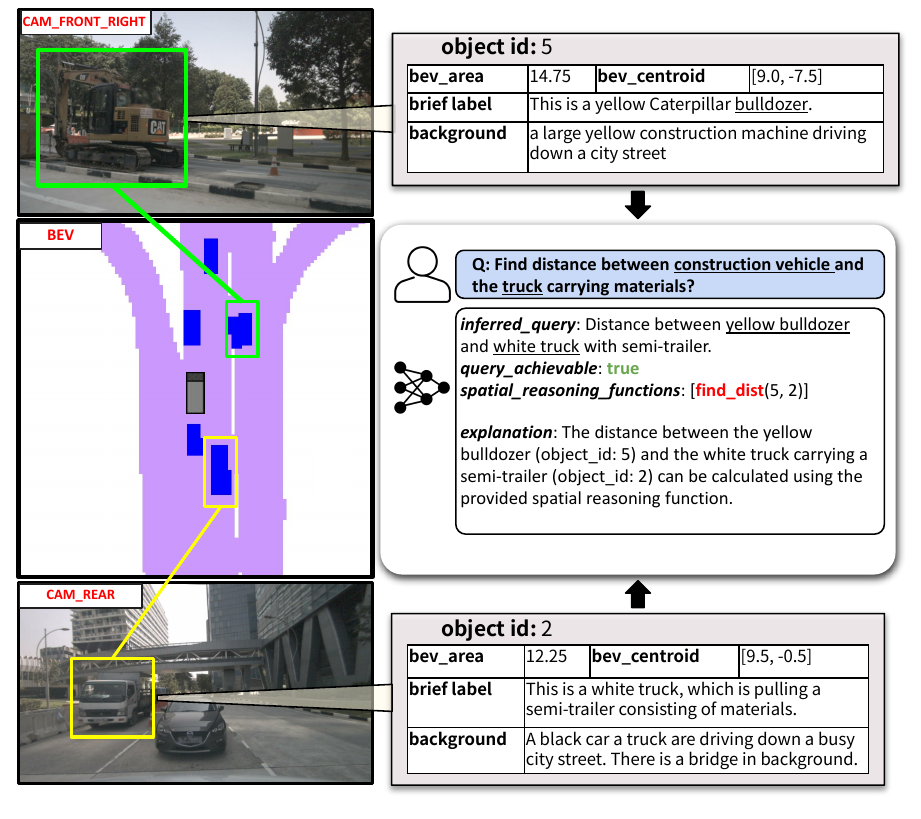}
    \caption{\textbf{Spatial Operators}: To compute distance between bulldozer and white truck, the Language Enhanced Maps for the objects are interpreted by an LLM like GPT4 to invoke relevant spatial operators in our framework with appropriate object IDs as arguments.\vspace{-15pt}}
    \label{fig:spatial-operators-usage}
\end{figure}

\noindent\textbf{Spatial Operators}
To enable the LLM to accuratey perform spatial reasoning, we provide access to an API of primitive spatial operators, following~\cite{jatavallabhula2023conceptfusion}.
Whenever a user query involves spatial reasoning (locations, distances, orientations), the model is instructed to generate API calls that directly invoke one of these spatial operators, rather than directly attempting to produce an output.
A full list of these spatial operators is provided in Table~\ref{sp-ops}.
An example usage of spatiatl operators is illustrated in Fig.~\ref{fig:spatial-operators-usage}, where we are able to capture the distance between the construction vehicle and the truck carrying materials. Importantly, these vehicles are never co-visible in the same camera, and require a BEV map for reasoning about them jointly.

\subsection{Implementation Details}

To generate BEV maps from multi-view images, we use the Lift-Splat-Shoot model~\cite{philion2020lift}. Each BEV is a $200 \times 200$ grid, where each cell has a resolution of $0.5 m$.
All our ground-truth BEV maps (used for evaluation) have the same resolution and grid dimensions.
We experiment with a number of LVLMs to compute vision-language features --  BLIP-2~\cite{li2023blip2}, MiniGPT-4~\cite{minigpt4} and InstructBLIP-2~\cite{dai2023instructblip}.
These features are later used as context to language decoder of LVLM to output object descriptions. 
For BLIP-2, we use the Flan5XXL~\cite{flan5xxl} language decoder and for InstructBLIP-2 and MiniGPT-4, we use the Vicuna-13b language decoder~\cite{vicuna2023}.
We use the default temperature value of 0.7 for LVLM for all experiments.
We perform inference on NVIDIA DGX A100.

\small

\begin{table}[!htb]
\renewcommand{\arraystretch}{1.5}

\begin{tabularx}{0.48\textwidth}{p{3.3cm}X}
\toprule
\textbf{Method} & \textbf{Description} \\
\midrule
\scriptsize{\texttt{front\_filter(objs)}}  & objects to the front \\
\scriptsize{\texttt{left\_filter(objs)}} & objects to the left\\
\scriptsize{\texttt{right\_filter(objs)}} & objects to the right\\
\scriptsize{\texttt{rear\_filter(objs)}} & objects to the rear\\
\scriptsize{\texttt{dist\_filter(objs, X)}} & objects within \textbf{``X"}m \\

\scriptsize{\texttt{k\_closest(objs, k)}} & k closest objects \\
\scriptsize{\texttt{k\_farthest(objs, k)}} & k farthest objects\\

\scriptsize{\texttt{objs\_in\_dist(objs, id, dist)}} & objects within distance \textbf{``dist"} to $o_{id}$\\
\scriptsize{\texttt{k\_closest\_to\_obj(objs, id, k)}} &  {k} closest objects to $o_{id}$\\
\scriptsize{\texttt{k\_farthest\_to\_obj(objs, id, k)}} & {k} farthest objects to $o_{id}$\\
\scriptsize{\texttt{obj\_distance(objs, id)} }&  distance (in m) to $o_{id}$\\

\scriptsize{\texttt{find\_dist(objs, id1, id2)}} & distance between 2 objects $o_{id1}$, $o_{id2}$\\
\bottomrule
\end{tabularx}
\caption{\textbf{List of spatial operators}:  Here \texttt{objs} is the list of objects in the BEV, $o_{id}$ refers to the object whose \texttt{object\_id} is $id$. Operators that do not take in \texttt{object\_id} as input operate on the ego-vehicle.
\vspace{-10pt}}
\label{sp-ops}
\end{table}
\normalsize

\section{The Talk2BEV-Bench Benchmark}
To evaluate the quality of our language-enhanced map and assess the spatial understanding and visual reasoning capabilities of our framework, we present \coolname-Bench -- the first benchmark for assessing LVLMs for autonomous driving applications.
We generate ground-truth language-enhanced maps for 1000 scenes from the NuScenes dataset~\cite{caesar2020nuscenes}, and more than 20,000 human-verified question-answer pairs in the SEEDBench~\cite{li2023seedbench} format\footnote{Each question has multiple answer choices, with one correct answer.}.
The questions evaluate understanding of object attributes, instance counting, visual reasoning, decision making, and spatial reasoning.
To generate the questions and responses, we first extract ground-truth BEV maps from the NuScenes dataset and obtain captions for each object in the map. The captions are refined by human annotators, after which we employ GPT-4 to generate questions and initial responses for each question. These questions and responses are, again, validated by human annotators to result in the final set of MCQs used in the benchmark.
This question and answer curation approach is illustrated in Fig.~\ref{fig:json_generation}, with an example set of generated questions given a ground-truth language-enchanced BEV map.

\subsection{Ground-truth language-enhanced maps}
We first use the BEV maps provided as part of the NuScenes ground-truth data to identify objects of interest, and obtain their image crops by LiDAR-camera projection.
For each object, we extract captions for its foreground and background context.

\noindent\textbf{Crop captions:} 
We employ a dense captioning model (GRiT~\cite{grit}) to generate text descriptions encapsulating fine-grained details within each object bounding box.
We also leverage an off-the-shelf text recognition model (PaddleOCR~\cite{du2020ppocr}), extracting any foreground text, to enhance understanding of object type and category.
    
\noindent\textbf{Background information:}
In addition to object-level (foreground) caption, we also extract information about the scene context (background) features by captioning the images.
This captures additional context such as street signs, barriers, weather conditions, time of day, and unique scene elements. Human annotators verify and refine the combined foreground and background captions at this stage, as shown in Fig.~\ref{fig:json_generation}.

\subsection{Question Generation and Evaluation Metrics}\label{subsec:eval}
Our evaluation spans four types of visual and spatial understanding tasks -- \emph{instance attributes} (questions pertaining to objects and their attributes), \emph{instance counting} (counting the number of objects that correspond to the text query), \emph{visual reasoning} (questions assessing general visual understanding questions not directly captured in the other categories), and \emph{spatial reasoning} (questions pertaining to location, distance, or orientation information).
For each scene and evaluation dimension, we prompt GPT-4 five times to generate five such questions per dimension, resulting in $20$ questions per scene.
For all categories (except spatial operators), we report an accuracy metric (since the questions are multiple-choice).
For spatial reasoning queries, we report regression metrics in the form of Jaccard index (for queries that expect a set of objects as output) and distance errors (for queries that require distance values as output).

\section{Results}

\begin{figure*}[!htbp]
    \centering
    \includegraphics[width=0.85\linewidth]{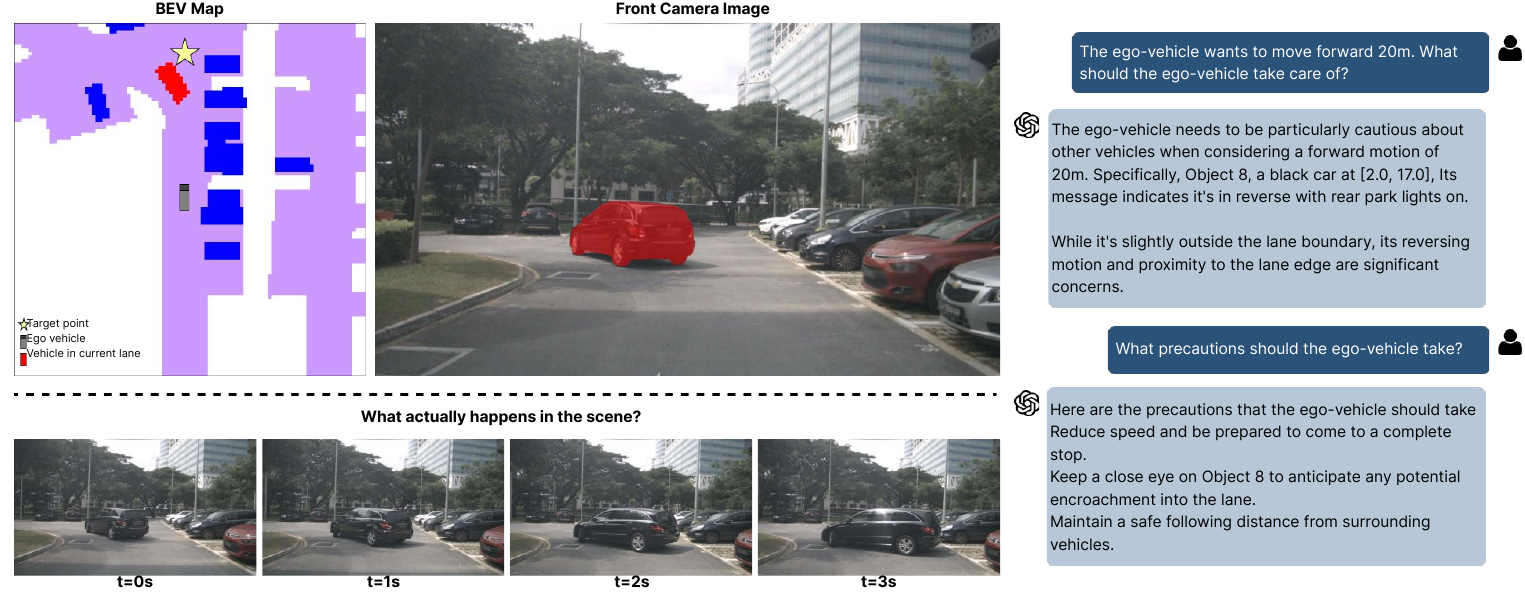}
    \caption{\coolname{} in \textbf{free-form conversation} with a user.
    There is a car in front of the ego-vehicle (highlighted in red), which is reversing to park in a parking spot. \coolname{} identifies the parking lights are on, and based on this visual information, and the spatial location of the car in front, \coolname{} deems it unsafe to continue moving forward.
    }
    \label{fig:free-form}
\end{figure*}

In this section, we evaluate \coolname{} quantiatively on questions from \coolname-Bench, and find that
\begin{enumerate}
    \item \coolname{} addresses a broad set of visual and spatial understanding tasks by leveraging the language-enhanced maps
    \item Access to an API of primitive spatial operators significantly improves performance on spatial reasoning tasks
    \item The zero-shot nature of \coolname{} allows seamlessly switching LVLMs, enabling easy integration across more performant LVLMs.
\end{enumerate}
We also present qualitative results on challenging scenarios from NuScenes~\cite{caesar2020nuscenes}, indicating the ability of \coolname{} to interpret the BEV layout at a granularity that allows predicting potential risky driving maneuvers and recourse.

\subsection{Quantitative Results}
We first asess the performance of \coolname{} on questions from \coolname-Bench. In Table~\ref{accuracy-LVLMs}, we report the performance across task subsets and across LVLMs used.
To delineate errors originating from incorrect BEV predictions versus inaccurate LVLM captions, we also present results from an oracle approach that leverages the ground-truth BEV map.
When using BEV maps output by LSS~\cite{philion2020lift}, we find that InstructBLIP-2 achieves the best performance in \textit{instance attribute} recognition and \textit{visual reasoning} compared to the BLIP-2 and MiniGPT-4 counterparts. In contrast, for \textit{instance counting}, MiniGPT-4 based $\textbf{L}(\mathcal{O})$  map achieves the best accuracy. 
Overall, we notice that MiniGPT-4 achieves best average performance across different types of questions.
 We notice that \textit{instance attribute} and \textit{visual reasoning} tasks are more sensitive to the quality of LVLM captions compared to other question categories, which is expected given the complexity of these tasks compared to \textit{instance counting}.
 We also note that errors in the BEV have only a minor impact on performance (3\%); meaning that as more performant LVLMs are released, the performance of \coolname{} is expected to improve further.
\begin{table}[!htbp]
\centering
\begin{tabular*}{0.49\textwidth}{@{\extracolsep{\fill}}ccccccc}
\toprule
\textbf{BEV} & \textbf{LVLM} & \thead{\textbf{Instance}\\\textbf{Attribute}} & \thead{\textbf{Instance}\\\textbf{Counting}} & \thead{\textbf{Visual}\\\textbf{Reasoning}} & \textbf{Avg} \\
\midrule
 & BLIP-2 & 0.50 & 0.83 & 0.47 & 0.60 \\
 LSS & InstructBLIP-2 & \textbf{0.54} & 0.80 & \textbf{0.50} & 0.62 \\
 & MiniGPT-4 & 0.50 & \textbf{0.90} & 0.49 & \textbf{0.63} \\
 \midrule
 & BLIP-2 & 0.51 & 0.83 & 0.47 & 0.60 \\
 GT & InstructBLIP-2 & \textbf{0.55} & 0.80 & 0.50 & 0.62 \\
 & MiniGPT-4 & \textbf{0.55} & \textbf{0.91} & \textbf{0.51} & \textbf{0.66} \\
\bottomrule
\end{tabular*}
\caption{\textbf{Overall Accuracy on MCQ Queries ($q_{mcq}$)}. Performance of \coolname{} with Language Enhanced Map constructed with different LVLMs (BLIP-2, InstructBLIP-2, MiniGPT-4) and BEV variants (LSS and GT)  on Multiple Choice Questions (MCQs).}
\label{accuracy-LVLMs}
\end{table}

\subsection{Qualitative Results}

\begin{figure*}[!htbp]
    \centering
    \includegraphics[width=0.95\linewidth]{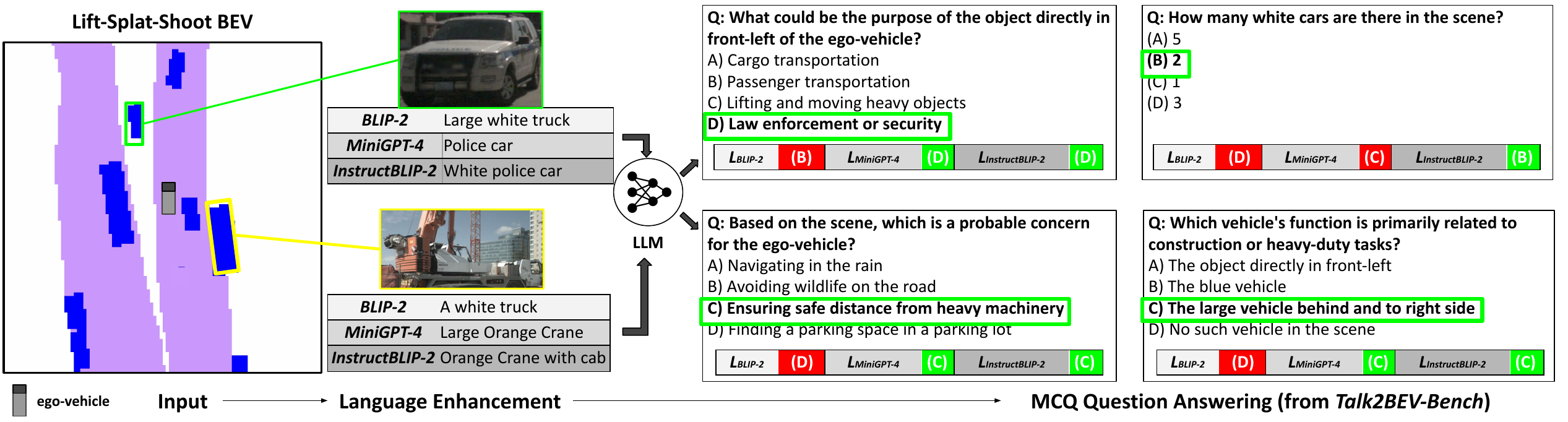}
    \caption{\textbf{Qualitative Results:} A BEV corresponding to a scene with multiple vehicles at an interchange.
    \coolname{} is able to identify emergency vehicles (such as the \emph{police car} shown here).
    The captions for a police car and a construction vehicle from Language Enhanced maps constructed with  with different LVLMs (BLIP-2, InstructBLIP-2, MiniGPT-4) have been visualized.
    We show the corresponding BEV captions produced by various LVLMs and their performance across 4 questions from \textit{Talk2BEV-Bench} relevant to these 2 objects.
    The correct answer for each question is highlighted in green.
    }
    \label{fig:qual}
\end{figure*}

In Fig.~\ref{fig:free-form}, we show a free-form interactive dialogue with \coolname{} where the user intends to advance by 20 m and inquires about potential obstructions. Ahead of the ego-vehicle is another vehicle reversing into a parking spot. \coolname{} leverages the vehicle's parking light and position information to deduce intent and advises caution. The LLM's prediction aligns with the vehicle's future activity from $t=0$ to $t=3$s. In Fig.~\ref{fig:qual}, we compare the performance of multiple LVLMs on MCQ queries from \coolname-Bench.

\subsection{Impact of Spatial Operators.}
\label{sec:results}

\begin{figure}
    \centering
    \includegraphics[width=\linewidth]{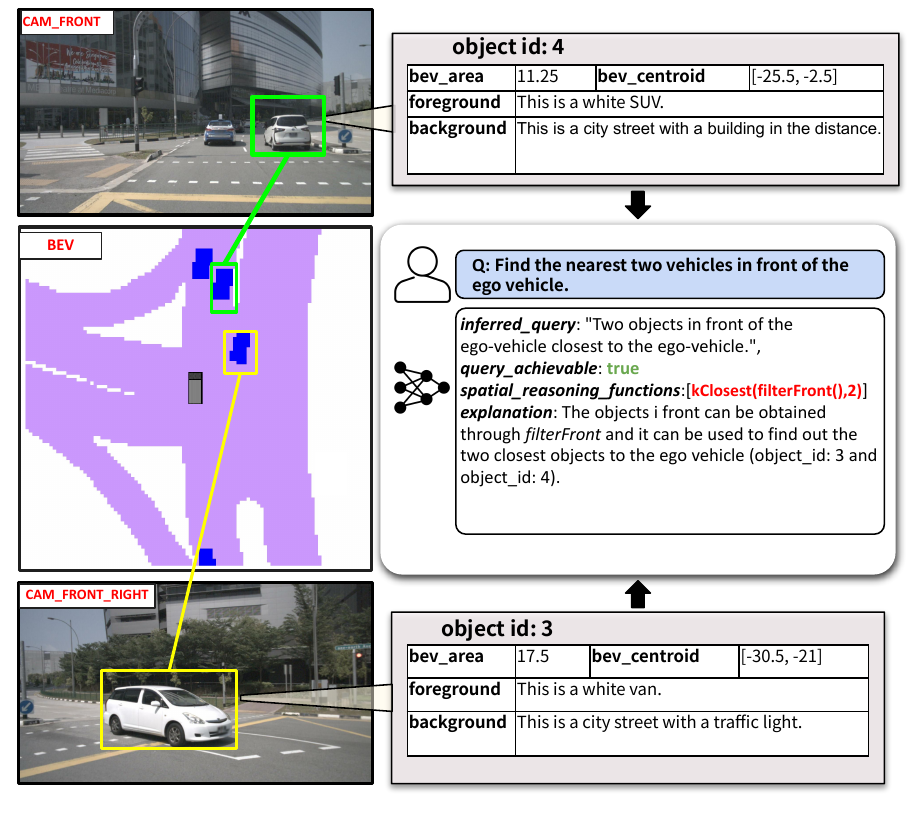}
    \caption{\textbf{Composition of Spatial Operators}: To find the nearest 2 vehicles in front, LLM like GPT4 composes the spatial operators.\vspace{-11pt} }
    \label{fig:spops-comp}
\end{figure}
\begin{table}[!htbp]
  \centering
  \begin{tabular}{lcccc}
    \toprule
    & \textbf{Jaccard Index $\uparrow$} & \textbf{Distance Error $\downarrow$} \\
    \midrule
    Random & 0.16 & 0.44 \\
    Talk2BEV w/o SO* & 0.25 & 0.22 \\
    Talk2BEV with SO* & \textbf{0.83} & \textbf{0.13} \\
    \bottomrule
    \addlinespace[0.5ex]
    *SO: Spatial Operators & & \\
  \end{tabular}  %
  \caption{\textbf{Impact of spatial operators}: When relying directly on the LLM's abilities to reason about distances, orientations, and areas, we notice a significant performance drop (\coolname{} w/o SO). Providing access to primitive spatial operators via API calls enables strong performance in terms of Jaccard index (higher is better) and distance error (lower is better) metrics. \vspace{-11pt}}
  \label{metrics-spops}
\end{table}

To assess the impact of explicit spatial operators available to our model via an API, we evaluate the performance of our system with and without spatial operators in Table~\ref{metrics-spops}.
Note that spatial reasoning queries are evaluated using Jaccard index or distance error based on nature of query as explained in Sec.~\ref{subsec:eval}.
For reference, we implement a baseline method, \emph{Random}, which uniformly randomly guesses distances and relevant objects.
While \coolname{} without spatial operators demonstrates markedly better performance compared to the \textit{Random} baseline, the model seems to struggle with spatial reasoning queries, often encountering large errors.
We see that \coolname{} integrated with our spatial operators achieves significant performance leaps (58\% improvement in Jaccard index, 0.09 m reduction in distance error) compared to directly using the LLM (here, GPT-4~\cite{openai2023gpt4}) for spatial reasoning.

\subsection{Performance across Object Categories}
To assess variance in performance across object categories, we report per-category statistics in Table~\ref{accuracy-types}.
We note that 2-Wheeler vehicles, including bicycles and motorcycles, consistently showed lower performance compared to other categories. This is mainly due to their smaller BEV segmentation predictions, making it more difficult to accurately back-project when there are minor inconsistencies in the predicted positions.
On the contrary, larger vehicles such as trucks and construction vehicles consistently outperformed cars in most cases.
This can be attributed to their larger BEV segmentations, which enable more accurate back projections.

\begin{table}[!htbp]
  \centering
\adjustbox{max width=\linewidth}{%
  \begin{tabular}{lccccc}
    \toprule
    \textbf{BEV} & \textbf{LVLM} & \textbf{2-Wheeler} & \textbf{Cars} & \textbf{Trucks} & \textbf{Construction} \\
    \midrule
    \multirow{4}{*}{LSS}
    & BLIP-2 & 0.56 & 0.60 & 0.67 & 0.67 \\
    & InstructBLIP-2 & 0.52 & 0.58 & 0.73 & 0.61 \\
    & MiniGPT-4 & 0.48 & 0.59 & 0.67 & 0.72 \\
    & \cellcolor{gray!25} \textit{Average}   & \cellcolor{gray!25} 0.52 & \cellcolor{gray!25} 0.59 & \cellcolor{gray!25} 0.69 & \cellcolor{gray!25} 0.67 \\
    \midrule
    \multirow{4}{*}{GT}
    & BLIP-2 & 0.56 & 0.60 & 0.68 & 0.67 \\
    & InstructBLIP-2 & 0.56 & 0.58 & 0.74 & 0.67 \\
    & MiniGPT-4 & 0.56 & 0.66 & 0.72 & 0.72 \\
    & \cellcolor{gray!25} \textit{Average} & \cellcolor{gray!25} 0.56 & \cellcolor{gray!25} 0.61 & \cellcolor{gray!25} 0.71 & \cellcolor{gray!25} 0.68 \\
    \bottomrule
  \end{tabular}
  } %
  \caption{\textbf{Object Category-wise Evaluation:} Performance of \coolname{} with Language Enhanced Map constructed with different LVLMs (BLIP-2, InstructBLIP-2, MiniGPT-4) and BEV variants (LSS and GT) on queries $q_{mcq}$ for different vehicle categories.\vspace{-5pt}}
  \label{accuracy-types}
\end{table}

\section{Conclusion}

In this work, we presented \coolname{}, a language interface to BEV maps used in autonomous driving systems.
By drawing upon recent advances in LLMs and LVLMs, \coolname{} caters to a variety of AD tasks, including, but not limited to, visual and spatial reasoning, predicting unsafe traffic interactions, and plotting recourse.
We also introduced Talk2BEV-Bench, a benchmark for evaluating subsequent work in LVLMs for AD applications.
While we continue to integrate large pretrained models into AD stacks, we also emphasize the need for safety and alignment research before these models are deployed into safety-critical AD stacks.

%%%%%%%%%%%%%%%%%%%%%%%%%%%%%%%%%%%%%%%%%%%%%%%%%%%%%%%%%%%%%%%%%%%%%%%%%%%%%%%%

% Bibliography
\bibliographystyle{IEEEtran}
\bibliography{IEEEtranBST/IEEEabrv, root}

\end{document}